\documentclass[11pt]{article}
\usepackage[margin=1.1in]{geometry}
\usepackage{graphicx}
\usepackage{lmodern}
\usepackage{microtype}
\usepackage{booktabs}
\usepackage{amsmath}
\usepackage[numbers,sort&compress]{natbib}
\usepackage{hyperref}
\usepackage{xurl}

\usepackage{caption}
\title{\fontsize{25pt}{45pt}\selectfont\textbf{Anterior’s Approach to Fairness Evaluation of Automated Prior Authorization System}}

\author{
    Sai P. Selvaraj \\
    \and
    Khadija Mahmoud, MD \\    
    \and
    Anuj Iravane
}\date{}

\begin{document}
\maketitle
\vspace{-1.2cm}
\begin{center}
{\fontsize{14pt}{17pt}\selectfont Anterior, Inc.}
\end{center}
\vspace{.5cm}

\begin{abstract}
\noindent Increasing staffing constraints and turnaround-time pressures in Prior authorization (PA) have led to increasing automation of decision systems to support PA review. Evaluating fairness in such systems poses unique challenges because legitimate clinical guidelines and medical necessity criteria often differ across demographic groups, making parity in approval rates an inappropriate fairness metric. We propose a fairness evaluation framework for prior authorization models based on model error rates rather than approval outcomes. Using 7,166 human-reviewed cases spanning 27 medical necessity guidelines, we assessed consistency in sex, age, race/ethnicity, and socioeconomic status. Our evaluation combined error-rate comparisons, tolerance-band analysis with a predefined $\pm$5 percentage-point margin, statistical power evaluation, and protocol-controlled logistic regression. Across most demographics, model error rates were consistent, and confidence intervals fell within the predefined tolerance band, indicating no meaningful performance differences. For race/ethnicity, point estimates remain small, but subgroup sample sizes were limited, resulting in wide confidence intervals and underpowered tests, with inconclusive evidence within the dataset we explored. These findings illustrate a rigorous and regulator-aligned approach to fairness evaluation in administrative healthcare AI systems.
\end{abstract}

\section{Introduction}
Prior authorization (PA) is a central administrative process used by health insurers or payers to determine whether a requested medical service meets medical necessity criteria according to a protocol\footnote{We use ``protocol" to refer to a medical necessity policy or guideline, that specifies the set of clinical criteria a requested service must meet to be approved for coverage by an insurer.} before coverage approval. As PA decision-support systems are increasingly deployed at scale, evaluating their fairness across demographic groups is a critical requirement for both internal governance and regulatory accountability. Systematic differences in model performance across demographic groups could lead to disproportionate delays or unnecessary escalations for certain populations, thereby directly affecting patient care and provider workflow.

However, fairness evaluation in prior authorization differs from many other clinical AI settings. Approval rates may vary across groups due to protocol-driven clinical differences, reflecting physiological and epidemiological variation observed across demographic groups. Symptoms, diagnostic thresholds, risk factors, and recommended treatment pathways may differ by sex, age, or other demographic characteristics, reflecting evidence-based clinical practice rather than inequitable treatment \citep{iom_guidelines_2011, uspstf_current_processes}. As a result, fairness assessments based solely on outcome parity risk conflate appropriate clinical variation with model bias. This conflation can lead to false conclusions—either incorrectly flagging appropriate clinical variation as bias or masking true model disparities behind population differences \cite{barocas-hardt-narayanan, obermeyer2019dissecting}.

To address this challenge, we frame fairness in prior authorization as a question of model performance consistency, rather than outcome parity. Specifically, we evaluate whether an automated PA model makes errors (compared to human reviewers) at different rates across demographic groups. Our primary fairness metric is the error rate, which isolates model behavior from baseline clinical variation and directly measures alignment with human decision-making, which is the relevant fairness concern in administrative, reversible workflows such as PA \cite{933006}.

Prior authorization decisions are based on applying payer-specific medical necessity guidelines to heterogeneous clinical documentation of a patient, both of which introduce substantial complexity. Patient clinical records are often 100s of pages long, containing multiple unstructured documents (e.g., progress notes, imaging reports, referral letters), and are frequently transmitted via fax or scanned uploads. Relevant information supporting the requested service may be distributed across documents, often requiring manual reconciliation. Decisions are based on payer-defined clinical guidelines (policies), which are often dozens of pages long (sometimes in the 100s) and encode detailed, condition-specific, hierarchical medical necessity rules \cite{cms_medicare_integrity_ch13, ama_policy_320_939} and may include multiple conditional pathways. This complexity in the data and decision make the task inherently complex.

Manual prior authorization review is both resource-intensive and time-sensitive, with strict turnaround-time requirements. Beyond clinical review, which itself can take tens of minutes per case for trained reviewers, substantial effort is expended on the manual, asynchronous administrative workflows that support the review, including fax intake, indexing, eligibility checks, and request validation. This is non-trivial, even for trained reviewers, and can have variability in decision quality \cite{caqhi_2022}. Delays can postpone diagnostic testing or treatment initiation, directly affecting patient care and provider workflow.

The volume and complexity of prior authorization requests have increased steadily year-over-year \cite{ama_pa_survey_2023}, while Medicare Advantage oversight and regulatory scrutiny have intensified \cite{oig_ma_prior_auth_2022}, making it difficult to scale. In response, payers are increasingly deploying automated and semi-automated decision-support systems to assist human reviewers by recommending approvals or escalating cases for manual review \cite{mckinsey_pa_ai_2022}.

In this work, we evaluate the bias of our automated PA workflow that operates over unstructured clinical documentation and apply payer-specific guidelines to produce a categorical recommendation. Our automation system uses contextualized LLMs and symbolic scaffolds to make automated decision suggestions using the inputs. While our PA systems may emit multiple outcomes, we restrict our analysis to a binary framing: approval versus escalation for human review\footnote{Our automation workflow does not produce or recommend denials, only recommends approval or escalation for human review.}, to simplify evaluation. Using a large corpus of real, human-reviewed prior authorization cases spanning multiple clinical protocols across various specialities, we present a multi-layered statistical methodology for bias testing based on model error rates. Our methodology combined unadjusted hypothesis testing, confidence-interval tolerance analysis, statistical power assessment, and adjusted regression controlling for guideline-level confounding. To our knowledge, this work represents one of the first systematic fairness evaluations of a prior authorization decision-support model grounded in real-world clinical operations and guideline-driven decision making.

\section{Related Work}

\subsection{Fairness Evaluation in Machine Learning and Healthcare}

A large body of work in machine learning defines group fairness through criteria such as demographic parity, equalized odds, and equality of opportunity \cite{hardt2016equality, dwork_fairness_2012}. These metrics are typically framed in terms of parity of outcomes or conditional error rates and assume that protected attributes should not influence decisions except through permissible pathways. However, prior work has shown that many fairness criteria are mutually incompatible and that metric choice must depend on the application context \cite{kleinberg2016inherent}. When base rates differ across groups, several authors argue that fairness should be evaluated using error rates rather than outcome rates \cite{chouldechova2017fair}. In contrast to approaches that treat outcome parity violations as evidence of bias, we explicitly reject approval-rate parity as a fairness objective in prior authorization and instead evaluate subgroup differences in model error rates.

Fairness evaluation in healthcare AI has primarily focused on predictive and diagnostic models, including disease detection, risk stratification, and clinical outcome prediction \cite{rajkomar2018ensuring, chen2021ethical}. Empirical studies show that models optimized for aggregate performance can exhibit disparities due to biased proxies, label bias, or uneven data coverage \cite{obermeyer2019dissecting}. At the same time, several authors caution that fairness in healthcare cannot be evaluated independently of clinical context, as observed differences in outcomes may reflect epidemiological variation, guideline-driven practice, or disease prevalence rather than algorithmic bias \cite{chen2019can, wiens2019no}.

Despite substantial work on fairness in clinical prediction and diagnostic AI, there is limited research on fairness evaluation in utilization management and administrative decision-support systems, such as prior authorization. These systems differ fundamentally from clinical prediction tasks: they operate under explicit payer-defined guidelines, incorporate routine human oversight, and primarily affect access and administrative burden rather than direct clinical outcomes. As a result, fairness frameworks developed for diagnostic or prognostic models do not directly transfer to prior authorization settings. We therefore evaluate fairness by examining subgroup differences in model error rates while conditioning on protocol variation, thereby providing a framework tailored to administrative healthcare AI systems.

\subsection{Approval-Rate Parity and Its Limitations}

Many applied fairness analyses compare approval or positive-decision rates across demographic groups \cite{feldman2015certifying}. While intuitive, this approach implicitly assumes that equal approval rates correspond to fair outcomes. In healthcare utilization and coverage decisions, however, approval rates are often determined by guideline eligibility, disease severity, and treatment appropriateness, all of which may legitimately vary across demographic groups \cite{graham2011institute}. As a result, enforcing approval-rate parity can mischaracterize appropriate clinical variation as algorithmic bias or incentivize clinically inappropriate decisions \cite{pmlr-v106-pfohl19a}. For this reason, we reject approval-rate parity as a fairness objective in prior authorization. Instead, we evaluate fairness by examining subgroup differences in model error rates conditional on review, thereby isolating model performance from guideline-driven variation in approval outcomes.

\subsection{Equivalence Testing and Tolerance-Based Evaluation}

In contrast to null-hypothesis significance testing, equivalence and non-inferiority testing frameworks are designed to determine whether observed differences are small enough to be practically irrelevant \cite{walker2011understanding}. These methods require explicit specification of an acceptability margin that reflects domain-specific judgment rather than statistical convention.
Several recent works and standards recommend applying tolerance-band and confidence-interval–based reasoning to algorithmic fairness, particularly in operational settings \cite{tatem2017comparing, steegen2016increasing}. This approach distinguishes statistically detectable differences from differences that are large enough to matter in practice.

We adopt this perspective directly. Rather than treating any statistically significant difference as evidence of bias, we evaluated whether confidence intervals for error-rate differences fall within a predefined operational tolerance band. We further assessed the width of confidence intervals relative to the tolerance threshold to avoid over-interpreting imprecise estimates. This goes beyond most prior fairness analyses, which report point estimates or p-values without explicitly reasoning about practical significance or statistical power.

\subsection{AI in Healthcare Workflow Automation}

Artificial intelligence is increasingly used to automate healthcare workflows beyond predictive modeling \cite{rajkomar2018scalable}, including clinical documentation \cite{Selvaraj2021, yao2023improving, ramprasad2023generating, patel2020weakly}, improving clinical communication \cite{selvaraj2025automating, PandiSelvarajA90}, risk stratification \cite{obermeyer2019dissecting}, and provider-facing decision support \cite{topol2019high, pandi2025patient}. However, while provider-facing and diagnostic applications have been extensively studied—including subgroup fairness evaluations \cite{rajkomar2018ensuring, chen2021ethical, chen2019can, wiens2019no, obermeyer2019dissecting}—payer-side administrative systems such as prior authorization have received comparatively less attention in the academic literature.




\subsection{Regulatory and Standards Guidance}

Regulatory and standards bodies increasingly emphasize context-aware fairness evaluation and caution against rigid parity requirements. National Institute of Standards and Technology (NIST) Special Publication (SP) 1270 explicitly recommends combining statistical testing with confidence intervals, power analysis, and domain-informed thresholds when assessing bias in AI systems \cite{933006}. Similar guidance in healthcare AI stresses the importance of aligning evaluation methods with system use, oversight, and harm pathways \cite{us2021good}.
Our methodology operationalizes these recommendations in the context of prior authorization. By focusing on error rates, incorporating tolerance bands, evaluating power, and controlling for protocol-level confounding, we implement a fairness evaluation strategy that is consistent with emerging regulatory expectations while remaining grounded in real-world payer operations.

\section{Methodology}
\label{section:methods}

We evaluated demographic fairness in a prior authorization decision-support model by assessing whether its model error rates differed meaningfully across protected groups. Rather than enforcing outcome parity, our methodology defined fairness in terms of consistency of model performance relative to human reviewers, while explicitly accounting for guideline-driven clinical variation, statistical uncertainty, and data limitations.

\subsection{Prior Authorization Workflow and Human Review Process}

Prior authorization case requests consisted of patient clinical documentation submitted by providers together with payer-defined medical-necessity guidelines (“protocols”). In Anterior's deployed prior-auth automation system, incoming clinical documentation and protocols are preprocessed and contextualized using symbolic scaffolds and large language models to support a protocol-driven medical necessity assessment. The system produces a preliminary determination recommendation—either approval or escalation for human review.

The cases used in this work, after the automated recommendation workflow, also underwent human review conducted by licensed registered nurses with multiple years of clinical experience. Reviewers were trained in utilization management and medical necessity determination and operated within standardized workflows. They used the same inputs as the automation system and made a determination for each case, approval or escalation for further review. For this evaluation, we compared the automation system’s outputs with human-reviewed determinations.

\subsection{Fairness Estimand}

The determinations of our trained human reviewers served as the operational reference standard for all analyses. For each case in this work, we compared the automated system’s recommendation with the human reviewer’s determination and assigned one of four outcomes: Correct Approval, Incorrect (False) Approval, Correct Escalation, or Unnecessary Escalation. We defined a binary error indicator for each case $i$ as:

$$
\text{error}_i =
\begin{cases}
1, & \text{if Review Outcome} \in {\text{\{False Approval, Unnecessary Escalation\}}} \\
0, & \text{if Review Outcome} \in {\text{\{Correct Approval, Correct Escalation\}}}
\end{cases}
$$

\noindent\textbf{Error rate disparity:} For a protected attribute ($A$) with groups ($g \in \{0,1\}$), the group-specific error rate over the reviewed cases was defined as:

$$
p_g = \Pr(\text{error} = 1 \mid A = g)
$$

The primary fairness estimand, the error-rate disparity, is $\Delta = p_1 - p_0$. Fairness evaluation consisted of determining whether $\Delta$ differed meaningfully from zero. This framing treats fairness as a property of model performance consistency, rather than as equality of approvals, and aligns with guidance that fairness metrics should reflect task context and harm pathways \cite{933006}.

\subsection{Observed Data and Assumptions}

\subsubsection{Dataset scope and freezing:}

Fairness evaluation was conducted on 7,166 human-reviewed prior authorization cases collected over a fixed evaluation window. Only cases that underwent human review were included, ensuring that all evaluated decisions were accompanied by an associated reference outcome. The analysis used a single frozen model version and a static dataset snapshot; no data were added, removed, or re-labeled during evaluation. For each case, we observed: (i) the protocol identifier, (ii) a reviewer adjudication label, and (iii) demographic attributes of the patient.

The dataset spanned multiple payer-defined clinical protocols across various specialties. Protocol identifiers were retained for all cases and were used explicitly in adjusted analyses; no protocol-level filtering was applied beyond standard data quality checks.

\subsubsection{Demographic attributes:}

The model operated on unstructured clinical documentation and did not receive canonical demographic metadata unless such information was explicitly present in the input clinical. Accordingly, demographic attributes used for fairness evaluation (sex, date of birth, race/ethnicity, payer line of business) were extracted solely from the clinical documentation. 
Attribute construction:
\begin{itemize}
    \item Sex: Standardized to Female and Male.
    \item Age group: Derived from date of birth and categorized as Adult (22–50) or Older Adult (51+). This threshold also aligns with established screening guidelines and healthcare utilization patterns \cite{smith2012use}”, as age-related disease prevalence shifts significantly around age 50 \cite{seer_csr_1975_2018, buttorff2017multiple, smith2012use}. Cases $\le 21$ years were excluded due to limited cases in the dataset.
    \item Race/ethnicity: Grouped into White vs Non-white based on explicit documentation.
    \item Socioeconomic status (SES): Proxied using payer line of business, mapping with Commercial and Employer-sponsored plans mapped to High SES and Medicaid and Marketplace plans mapped to Low SES, similar to mapping by \cite{seyyed2021underdiagnosis, monuteaux2024evaluation}. Medicare and other non-SES categories (such as auto accident, self-pay, and other) were excluded as they do not provide any economic marker.
\end{itemize}

Demographic attributes were extracted from the unstructured clinical documentation only when explicitly present. When a protected attribute was absent or not extractable, cases were excluded from analyses involving that attribute. For covariates used solely for adjustment, missing extracted values were encoded as an explicit ``Unknown'' category to preserve sample size. Our approach reflected that the model had access only to information explicitly documented in the clinical note and avoided imputing demographic metadata. This design reflected the information actually visible to the model and followed best practices for bias evaluation under partial observability \cite{933006}.

The analysis was conducted using real-world, post-deployment data (with no protected health information (PHI)) from a prior authorization decision-support system operating within routine clinical workflows.

\subsection{Null Hypothesis Testing}
\label{section:methods_null_hypo}

As an initial screening step, we performed null hypothesis testing by computing unadjusted model error rates for each protected group and conducting a two-proportion z-test to evaluate whether any statistically detectable disparity exists. For a protected attribute $A \in \{0,1\}$, we define the group-specific error rate and compute the unadjusted disparity (absolute difference in error rates) as:

$$
\widehat{p}_g = \frac{\sum_{i: A_i=g} \text{error}_i}{\sum_{i: A_i=g} 1} \quad \text{and} \quad \widehat{\Delta} = |\widehat{p}_1 - \widehat{p}_0|.
$$

We then formally test the two hypotheses
$H_0: \Delta = 0$ vs $H_1: \Delta \ne 0$. Under $H_0$, the two groups are assumed to have equal underlying error rates. And, under the alternative hypothesis $H_1$, the error rates are not constrained to be equal.

To evaluate $H_0$, we conducted a two-sided two-proportion z-test comparing the unadjusted group error rates. The z-statistic was computed using the standard normal approximation for independent binomial proportions, and corresponding two-sided p-values were derived from the standard normal distribution. Statistical significance was assessed at $\alpha = 0.05$. A p-value below $\alpha = 0.05$ indicates statistical evidence against the null hypothesis.

\paragraph{Limitations of NHST for Fairness Evaluation:}

While null hypothesis testing detects whether a statistically significant difference exists, it does not quantify whether an observed difference is practically meaningful. In large samples, small, operationally irrelevant differences may become statistically significant. Conversely, in smaller subgroups, meaningful disparities may fail to reach significance due to limited power.

Importantly, failure to reject $H_0$ (i.e., $p \ge 0.05$) does not imply that the groups are similar, only that insufficient evidence exists to detect a difference. For fairness evaluation, particularly in administrative healthcare workflows, the key question is not whether the difference is exactly zero, but whether it exceeds a predefined threshold of operational relevance.

For these reasons, null hypothesis testing is used only as a preliminary screening tool. Our primary fairness assessment relies on confidence-interval–based evaluation relative to a predefined tolerance band, described in the following section.

\subsection{Practical Significance via Tolerance Bands}
\label{section:methods_unadjusted}

To assess whether observed disparities were operationally meaningful, we evaluated a 95\% Wald confidence interval for the difference in error rates ($\Delta$) between groups, $[CI_{min}, \ CI_{max}]$, against a predefined tolerance band $[-\delta, +\delta]$:

\begin{itemize}
    \item If the CI lies entirely within the tolerance band — $\{CI_{min} \ge -\delta,\  CI_{max} \le +\delta\}$
    \begin{itemize}
        \item Any true difference, if present, was operationally negligible
        \item No meaningful disparity was identified (equivalence)
    \end{itemize}
    \item If the CI fully contained —
    $\{CI_{min} < -\delta,\ CI_{max} > +\delta\}$, or partially overlapped the tolerance band — $\{CI_{min} < -\delta,\ CI_{max} \le +\delta\}$, or $\{CI_{min} \ge -\delta,\ CI_{max} > +\delta\}$
    \begin{itemize}
        \item Both negligible and meaningful disparities were plausible given the data.
        \item The data was inconclusive.
    \end{itemize}
    \item  If the CI did not overlap the tolerance band —  $\{CI_{min} > +\delta\}$ or $\{CI_{max} < -\delta\}$
    \begin{itemize}
        \item  An operationally significant difference was present
        \item  A meaningful disparity existed (non-equivalence)
    \end{itemize}
\end{itemize}

In our work, we set the tolerance band to $\delta=\pm$5 percentage points (pp), reflecting the smallest difference considered material in a prior authorization workflow. This approach followed equivalence-style reasoning, which emphasized practical significance over purely statistical differences \cite{walker2011understanding}. The analyses use pre-specified parameters held constant across comparisons: significance level of $\alpha = 0.05$, $95\%$ confidence intervals, and a tolerance band of $\delta=\pm$5pp.

We additionally report the Evidence Ratio, defined as the width of the confidence interval relative to the tolerance span,
$\text{Evi. Ratio} = (CI_{max}-CI_{min})/{2\delta}$, to characterize evidentiary strength (strong, moderate, or weak), and to ensure that conclusions were not drawn from imprecise estimates.

\subsection{Power Analysis}
\label{section:methods_power_analysis}

Because subgroup sizes varied substantially across attributes, we computed achieved statistical power for each unadjusted comparison. Power was defined as the probability of detecting a disparity of magnitude ($\delta$) at a significance level of $\alpha$ = 0.05.
We treat $power \ge 0.80$ as sufficient to support conclusions regarding the absence of meaningful disparity. Results with lower power were interpreted as inconclusive, consistent with guidance for subgroup and equivalence analyses \cite{walker2011understanding, chow2017sample}.

\subsection{Adjusted Estimation with Protocol Control}
\label{section:methods_adjusted}
Unadjusted comparisons in Section \ref{section:methods_unadjusted} could be confounded by protocol mix, as different guidelines exhibited different baseline error rates. To isolate demographic effects from guideline variation, we performed an adjusted analysis using logistic regression with protocol fixed effects. Adjusted group error rates were estimated via marginal standardization, and the adjusted error-rate difference was computed as the difference between these standardized rates. Uncertainty in the adjusted difference was quantified using bootstrap 95\% confidence intervals, obtained by resampling individual cases with replacement at the case level using a fixed random seed for reproducibility.
\noindent Adjusted group error rates were computed as:

$$
\text{error} \sim A + C(\text{Protocol}) + C(\text{Covariates})
$$

\noindent where, $A$ is the protected attribute of interest, $C(Protocol)$ denotes protocol fixed effects, and $C(Covariates)$ includes additional demographic variables when available. Equivalently, the model was written in probabilistic form as:

$$
\text{logit}\left(\Pr(\text{error}=1)\right)
= \beta_0 + \beta_A A + \gamma_{\text{protocol}} + w^\top \text{Covariates}
$$

\noindent where $\beta_0$ is the intercept, $\beta_A$ is the coefficient associated with the protected attribute,  $\gamma_{protocol}$ represents a vector of protocol-specific fixed effects and $w$ denotes coefficients for additional covariates.

As in the unadjusted analysis in Section \ref{section:methods_unadjusted}, adjusted confidence intervals (and precision diagnostics ) were evaluated against the tolerance band ($\delta$) to determine whether any observed disparity was operationally meaningful. Uncertainty was quantified using bootstrap 95\% confidence intervals, which are robust to subgroup imbalance and non-normality \cite{efron_tibshirani_1993}. The analysis used a significance level of $\alpha$ = 0.05, 95\% confidence intervals, and a tolerance band of $\delta=\pm$5pp. All statistical analyses are implemented using standard scientific computing libraries. Bootstrap confidence intervals were computed by 1000 resamples, with individual cases resampled with replacement at the case level, using a fixed random seed to ensure reproducibility.

For covariates in adjusted analyses, missing extracted values are represented as an explicit ``Unknown'' category to retain cases and to reflect the absence of documented demographic information in the submitted text. As a robustness check, we also reported complete-case analyses excluding cases with missing covariate values (results provided in Appendix \ref{App:C}).

\subsection{Single-Factor Stratified Analyses (Sensitivity Analysis)}
\label{section:methods_sensitivity}

In addition to the primary analyses described above, we conducted single-factor stratified logistic regressions as a sensitivity analysis. These models included the protected attribute and one control variable at a time (e.g., protocol, age, race, or SES), but did not include the full set of covariates simultaneously.

These analyses were not intended to serve as final fairness determinations. Instead, they were used to (1) assess whether any observed disparities were driven by a single confounding factor, (2) evaluate the robustness of the primary findings, and (3) aid interpretation of the adjusted results. Results from these analyses were provided in Appendix \ref{App:B} and \ref{App:C}.

\subsection{Scope and Non-Applicability}

The PA system produced deterministic binary outputs and did not emit calibrated probability scores. As a result, calibration-based subgroup fairness metrics (e.g., expected calibration error) were not applicable in this setting. Evaluation was strictly post-deployment and observational.

\noindent All fairness claims in this work were conditional on:
\begin{itemize}
    \item The reviewed-case population,
    \item The availability of extracted demographic signals in documentation, and
    \item The observed distribution of protocols.
\end{itemize}

\section{Results}

We report fairness evaluation results for four protected attributes (sex, age, race/ethnicity, and socioeconomic status (SES)) using unadjusted and adjusted error-rate comparisons. Results are presented in terms of absolute error-rate differences, confidence intervals, statistical power, and tolerance-band evaluations.

\subsection{Cohort Characteristics and Coverage}

Fairness evaluation is conducted on 7,166 human-reviewed prior authorization cases, spanning multiple payer-defined protocols (medical necessity guidelines). Only reviewed cases are included to ensure a consistent human reference standard. After attribute-specific filtering and standardization, final cohort sizes vary substantially:
\begin{itemize}
    \item Sex: 6,505 cases
    \item Age: 6,786 cases
    \item Race/Ethnicity: 920 cases
    \item SES (payer-based): 1,929 cases
\end{itemize}

The dataset includes a broad set of payer protocols, which are explicitly retained and used as fixed effects in adjusted analyses. Sufficient protocol-level overlap is observed for all the protected categories, see Appendix \ref{App:A}. For race, limited representation reduces the effective sample size. Additional dataset characterization and protocol-overlap diagnostics are reported in the Appendix \ref{App:A}.

\subsection{Unadjusted Error-Rate Disparities}
\label{section:result_unadjusted}

\begin{table}[!ht]
\centering
\caption{ $|$ \textbf{Unadjusted evaluation.} Subgroup error-rate comparisons and unadjusted fairness evaluation across protected attributes. The table reports group error rates, absolute error-rate difference ($\Delta$), z-statistics, and p-values for NHST, 95\% confidence intervals (CI) for error rate difference, inclusion within a predefined tolerance band ($\delta$ = $\pm$5pp), and statistical power for detecting $\delta$ with $\alpha=0.05$. All observed disparities are small, non-significant, and the CI remains within the tolerance threshold. Attribute groups: Female (F), Male (M), Adult (A), Older (O), Non-white (NW), White (W), Low SES (L), High SES (H)}

\label{tab:table1}
\begin{tabular}{l l c c c c c c}
\toprule
\textbf{Attr.} & \textbf{Groups} & \textbf{Error rates} & $\boldsymbol{\Delta}$ & \textbf{NHST (z, p)} & \textbf{95\% CI} & \textbf{CI $\boldsymbol{\in \delta}$} & \textbf{Power} \\
\midrule
\textbf{Sex} & F, M            & 5.8\%, 6.3\%     & $-0.5$ & $-0.91,\ 0.364$ & $[-1.74, 0.65]$ & Yes & 1.00 \\
\textbf{Age} & A, O   & 5.8\%, 6.3\%     & $-0.5$ & $-0.79,\ 0.429$ & $[-1.69, 0.71]$ & Yes & 1.00 \\
\textbf{Race} & NW, W & 6.7\%, 7.3\%     & $-0.6$ & $-0.26,\ 0.798$ & $[-4.82, 3.68]$ & Yes & 0.51 \\
\textbf{SES} & L, H        & 5.9\%, 5.7\%    & $+0.1$ & $+0.10,\ 0.920$  & $[-2.55, 2.83]$ & Yes & 0.88 \\
\bottomrule
\end{tabular}
\end{table}

Table \ref{tab:table1} summarizes unadjusted group-wise error rates and disparity estimates for each protected attribute. Across all attributes, unadjusted error-rate differences are small ($\Delta\le$   1pp). Two-proportion hypothesis tests yield no statistically significant differences for any attribute (all $p > 0.05$); however, as discussed in Section \ref{section:methods_null_hypo}, statistical significance is not treated as a definitive fairness criterion.

For sex, age, and SES, confidence intervals are narrow relative to the $\delta=\pm$5pp tolerance band, and achieved power exceeds the 0.80 threshold, indicating adequate sensitivity to detect operationally meaningful disparities. For race, the confidence interval remains within tolerance, but is substantially wider, and achieved power is limited, reflecting reduced subgroup representation rather than evidence of parity.

\subsection{Adjusted Error-Rate Disparities (Protocol-Controlled)}
\label{section:results_adjusted}

\begin{table}[!ht]
\centering
\caption{$|$ \textbf{Adjusted evaluation.} Adjusted error-rate disparities across protected attributes after controlling for protocol and additional demographic covariates. The table reports adjusted risk differences (percentage points), 95\% bootstrap confidence intervals (CI), inclusion within the predefined operational tolerance band ($\delta=\pm$5pp), the Evidence Ratio (CI width relative to $\delta$), and qualitative evidence strength. Across sex, age, and SES, adjusted disparities are small and fully contained within the tolerance threshold with strong to moderate precision; race-adjusted comparisons exhibit wider intervals due to limited sample size and are therefore less conclusive}
\label{tab:table2_adjusted_disparities}

\begin{tabular}{l l c c c l}
\toprule
\textbf{Attr.} & \textbf{Adj. Set} & \textbf{Adj.} $\boldsymbol{\Delta}$ & \textbf{95\% CI} & \textbf{CI $\boldsymbol{\in \delta}$} & \textbf{Evi. Ratio (Strength)} \\
\midrule

\textbf{Sex} 
  & Protocol + Race & $+0.52$ & $[-0.74, 1.69]$ & Yes & 24\% (Strong evidence) \\
  & Protocol + Age  & $+0.51$ & $[-0.64, 1.57]$ & Yes & 22\% (Strong evidence) \\
  & Protocol + SES  & $+0.16$ & $[-1.11, 1.51]$ & Yes & 26\% (Strong evidence) \\[4pt]

\textbf{Race} 
  & Protocol + Sex  & $+0.28$ & $[-4.72, 5.50]$ & No & 102\% (Weak evidence) \\
  & Protocol + Age  & $+0.11$ & $[-5.01, 5.69]$ & No & 107\% (Weak evidence) \\
  & Protocol + SES  & $-0.14$ & $[-5.79, 4.67]$ & No & 105\% (Weak evidence) \\[4pt]

\textbf{Age} 
  & Protocol + Sex  & $+0.26$ & $[-1.05, 1.45]$ & Yes & 25\% (Strong evidence) \\
  & Protocol + Race & $+0.30$ & $[-0.93, 1.57]$ & Yes & 25\% (Strong evidence) \\
  & Protocol + SES  & $+0.16$ & $[-1.24, 1.36]$ & Yes & 26\% (Strong evidence) \\[4pt]

\textbf{SES} 
  & Protocol + Sex  & $-0.39$ & $[-2.79, 2.69]$ & Yes & 55\% (Moderate evidence) \\
  & Protocol + Race & $-0.30$ & $[-2.60, 2.31]$ & Yes & 49\% (Strong evidence) \\
  & Protocol + Age  & $-0.57$ & $[-3.02, 2.28]$ & Yes & 53\% (Moderate evidence) \\

\bottomrule
\end{tabular}
\end{table}

To control for protocol-level confounding, we estimate adjusted error-rate disparities using logistic regression with protocol fixed effects, as shown in Table \ref{tab:table2_adjusted_disparities}. Adjusted group differences are computed via marginal standardization, with uncertainty quantified using bootstrap confidence intervals as described in Section \ref{section:methods_power_analysis}.

For sex, age, and SES, adjusted disparities remain well below the tolerance threshold band ($\delta=\pm5$pp) across all specifications, with confidence intervals that are narrow to moderate relative to the tolerance span (Evidence Ratio). These results are consistent with the unadjusted findings and indicate stable model performance across these attributes after accounting for protocol differences.

For race, adjusted estimates exhibit substantially wider confidence intervals, and some specifications yield confidence intervals that overlap or extend beyond the tolerance boundary. This reflects limited subgroup coverage, reduced statistical precision (and the lower power observed in Section \ref{section:result_unadjusted}), rather than consistent directional disparities.

As a robustness check, we repeated adjusted analyses using complete-case covariates (excluding ``Unknown'' levels). For sex, age, and SES, point estimates and tolerance-band conclusions were unchanged, with minor improvements in CI precision in some specifications. For models including race as a covariate, conclusions were more sensitive to missingness due to reduced effective sample size: the sex model became more precise (moderate→strong), while age and SES models exhibited wider uncertainty and occasionally crossed the $\pm5$pp tolerance boundary (see Appendix \ref{App:C}).

\subsection{Sensitivity Analyses: Single-Factor Stratification}
\label{section:results_sensitivity}
As a robustness check, we conducted single-factor stratified logistic regressions including the protected attribute and one control variable at a time (e.g., protocol, age group, race, or SES). These analyses yielded disparity estimates that were consistent in direction and magnitude with those from the fully adjusted models. Where discrepancies arose, they were attributable to reduced effective sample size and limited protocol overlap rather than systematic changes in estimated disparities. Full results are provided in Appendix \ref{App:B}.

\section{Discussion}
\subsection{Principal Findings}

In this study, we evaluated demographic fairness in a prior authorization decision-support model using an error-based fairness framework that evaluates consistency of model performance relative to human reviewers. Across protected attributes with sufficient data coverage - sex, age, and socioeconomic status (SES) - both unadjusted and protocol-controlled analyses show that differences in model error rates are small and well within the pre-specified $\pm5$ percentage point (pp) tolerance band. Observed differences are generally within $\pm1$pp, a magnitude consistent with normal operational variation.

For these attributes, confidence intervals are narrow relative to the tolerance span, and the achieved statistical power is sufficient to detect operationally meaningful disparities. Together, these results indicate no meaningful performance disparities across sex, age, and SES within the reviewed-case population.
For race/ethnicity, point estimates of error-rate differences are similarly small, but confidence intervals are wide, and achieved power is limited due to sparse subgroup representation. As a result, findings for race are inconclusive rather than confirmatory and reflect limited statistical precision rather than evidence of systematic disparity.

Overall, these results suggest that model performance is stable across most demographic dimensions when fairness is evaluated using error-based metrics aligned with the clinical and operational context of prior authorization. To our knowledge, this is the first fairness analysis of prior authorization systems that jointly evaluates subgroup error rates using protocol-aware stratification.

\subsection{Strengths and Limitations}

This study introduces a fairness evaluation framework tailored to prior authorization workflows. A key methodological strength is the use of model error rates, rather than approval or denial rates, as the primary fairness metric. In prior authorization, approval rates may legitimately differ across populations due to medical necessity criteria and disease prevalence. By evaluating disagreement between the model and trained human reviewers, the analysis isolates potential disparities in model behavior rather than differences driven by underlying clinical variation.

A second strength is the use of protocol fixed effects to account for heterogeneity in the protocols, as prior authorization decisions are governed by payer-defined protocols that vary substantially across clinical domains. Controlling for protocol ensures that comparisons between demographic groups are made within the same guideline context rather than across heterogeneous clinical criteria, improving interpretability and reducing the risk of spurious disparities driven by protocol mix.

Third, fairness conclusions are grounded in a confidence-interval–plus–tolerance-band framework rather than statistical significance alone. Evaluating confidence intervals against a predefined tolerance band enables the analysis to distinguish statistically detectable differences from operationally meaningful ones. Reporting achieved power and evidence strength further improves transparency and guards against over-interpretation of imprecise estimates.

Several limitations should also be noted. The analysis relies on human reviewer adjudications as the operational reference standard, which may itself contain variability or bias. Demographic attributes are extracted from clinical documentation, resulting in substantial missingness for some variables, particularly race and socioeconomic status. As a result, fairness conclusions are conditional on cases where demographic attributes are explicitly documented and therefore observable to the model.

In addition, demographic attributes were operationalized using coarse groupings to ensure adequate subgroup sizes and stable estimation. These simplified categories may obscure within-group variation or intersectional effects. In particular, race and ethnicity were collapsed due to limited documentation, and payer-based SES serves only as a proxy for socioeconomic position. Finally, analyses involving race are constrained by limited subgroup representation, resulting in reduced statistical power and wider confidence intervals. These limitations reflect data availability rather than deficiencies in the evaluation framework itself.

\subsection{Implications for Fairness Evaluation}

These findings highlight several considerations for evaluating fairness in operational healthcare AI systems. First, fairness assessments should prioritize performance-consistency metrics that reflect actual decision errors, rather than enforcing outcome parity, which may conflict with clinical guidelines. In prior authorization workflows, approval rates are often driven by medical necessity requirements and disease severity, making approval-rate parity an unreliable indicator of fairness. Evaluating subgroup differences in model error rates provides a more direct assessment of whether the system behaves differently across populations.

Second, fairness evaluations must explicitly account for data sufficiency and subgroup representation. When demographic coverage is limited, analyses should report uncertainty and avoid definitive claims about fairness or disparity. Improved demographic signal capture and documentation practices will be important for enabling more reliable fairness assessments, particularly for attributes such as race and ethnicity.

Third, tolerance-band–based evaluation frameworks provide a practical approach for operational fairness monitoring. By defining acceptable margins in advance and interpreting confidence intervals relative to these thresholds, organizations can distinguish between negligible variation and disparities that may warrant investigation. Pairing tolerance thresholds with confidence intervals and power analysis enables transparent and defensible fairness determinations in real-world deployments.

Finally, as prior authorization systems become increasingly automated, fairness monitoring should be treated as an ongoing governance process rather than a one-time validation exercise. Because PA decisions can affect access to care and administrative burden, routine and independent fairness audits will be essential to ensure that automated decision-support systems do not introduce systematic performance differences across demographic groups over time.

\section{Conclusion}

This study presents a context-aware framework for evaluating demographic fairness in prior authorization decision-support models. Our findings demonstrate that an error-based fairness framework, combined with protocol-aware adjustment, provides a more appropriate and interpretable assessment of fairness in prior authorization systems.
By focusing on model error rates, incorporating protocol-level adjustments, and grounding conclusions in tolerance-band and confidence interval–based reasoning, we provide a fairness evaluation approach aligned with the operational realities of prior authorization workflows.

Across sex, age, and socioeconomic status, analyses show small differences in model error rates with narrow confidence intervals relative to the predefined tolerance band, indicating no meaningful performance disparities within the reviewed-case population. For race and ethnicity, results remain inconclusive due to limited subgroup representation, highlighting the importance of data sufficiency when interpreting fairness analyses.

More broadly, the framework demonstrates how fairness evaluation in administrative healthcare AI systems can be grounded in task-specific metrics, transparent statistical reasoning, and explicit reporting of uncertainty. These principles provide a practical foundation for responsible deployment and ongoing oversight of automated prior authorization systems.

\bibliographystyle{unsrtnat}
\bibliography{ref}

\newpage
\appendix
\section{Appendix}
\subsection{Data Distribution and Cohort Characteristics}
\label{App:A}

Table \ref{tab:A1_demographics} reports the distribution of reviewed prior authorization cases by protected attribute. Percentages are computed independently for each attribute. Demographic attributes are extracted from unstructured clinical documentation only when explicitly present. Missingness, therefore, reflects documentation practices in prior authorization submissions rather than post hoc exclusion or data loss. The quality of the extractions was also assessed and found satisfactory.

Sex and age are observed for the majority of cases, supporting well-powered fairness analyses for these attributes. In contrast, race/ethnicity and socioeconomic status (SES) exhibit substantial missingness, reflecting the fact that such information is rarely explicitly documented in prior authorization submissions.

\begin{table}[!ht]
\centering
\caption{ $|$ \textbf{Demographic characteristics.} Distribution of protected attributes extracted from clinical in the reviewed prior authorization cases ($N = 7{,}166$)}
\label{tab:A1_demographics}
\begin{tabular}{l l r}
\toprule
\textbf{Attr.} & \textbf{Category} & \textbf{Frequency (\%)} \\
\midrule

\textbf{Sex}
  & Female  & 3{,}961 (55.3\%) \\
  & Male    & 2{,}544 (35.5\%) \\
  & Ignored & 0 (0.0\%) \\
  & Missing & 661 (9.2\%) \\[4pt]

\textbf{Age group}
  & 22--50 (Adult)  & 2{,}241 (31.3\%) \\
  & 51+ (Older)    & 4{,}545 (63.4\%) \\
  & Ignored & 248 (3.5\%) \\
  & Missing & 132 (1.8\%) \\[4pt]

\textbf{Race / Ethnicity}
  & White      & 757 (10.6\%) \\
  & Non-White  & 164 (2.3\%) \\
  & Ignored & 19 (0.3\%) \\
  & Missing    & 6{,}226 (86.9\%) \\[4pt]

\textbf{SES (Payer-based)}
  & High SES    & 1{,}571 (21.9\%) \\
  & Low SES    & 358 (5.0\%) \\
  & Ignored & 566 (7.9\%) \\
  & Missing & 4,671 (65.2\%) \\

\bottomrule
\end{tabular}
\end{table}

\begin{table}[!ht]
\centering
\caption{ $|$ \textbf{Protocol distribution.} Aggregated payer protocol representation in the reviewed prior authorization cases ($N = 7{,}166$)}
\label{tab:A2_protocols}
\begin{tabular}{l r}
\toprule
\textbf{Protocol Characteristic} & \textbf{Value} \\
\midrule
Number of distinct protocols & 27 \\
Median cases per protocol (IQR) & 99 (63--339) \\
Percentage of cases in top 5 protocols & 62.8\% \\
\bottomrule
\end{tabular}
\end{table}

To characterize the diversity of the protocols while preserving confidentiality, protocol-level statistics are reported in aggregated form in Table \ref{tab:A2_protocols}. Case volume is concentrated among a subset of frequently used protocols, reflecting typical utilization patterns in prior authorization workflows. 

To assess the identifiability of protocol-adjusted fairness analyses, we evaluated protocol-level overlap across protected attributes. We define protocol overlap as the presence of at least one case from each comparison group within a protocol, excluding missing or ignored attribute values. Only protocols with at least one eligible case are considered valid for a given attribute. Table \ref{tab:A3_protocol_overlap} summarizes the number of protocols exhibiting overlap for each protected attribute. For all protected attributes, high overlap rates indicate that most protocols include representation from both comparison groups, supporting the identifiability of within-protocol comparisons in adjusted analyses.

\begin{table}[!ht]
\centering
\caption{ $|$ \textbf{Protocol overlap.} The table reports the number of protocols containing at least one case from each comparison group for a given attribute after excluding missing values. High overlap rates indicate that most protocols include representation from both groups, supporting identifiability of within-protocol comparisons in adjusted analyses.}
\label{tab:A3_protocol_overlap}
\begin{tabular}{l c c c}
\toprule
\textbf{Attr.} & \textbf{Protocols with Overlap} & \textbf{Valid Protocols} & \textbf{Overlap Rate} \\
\midrule
Sex                  & 27 & 27 & 100.0\% \\
Age group            & 25 & 27 & 92.6\% \\
Race / Ethnicity     & 21 & 24 & 87.5\% \\
SES / Payer    & 23 & 26 & 88.5\% \\
\bottomrule
\end{tabular}
\end{table}

Table \ref{tab:A4_reviewer_outcomes} summarizes the distribution of reviewer adjudication outcomes used to define model error. Overall error rates are low and balanced across false approvals and unnecessary escalations.

\begin{table}[!ht]
\centering
\caption{ $|$ \textbf{Outcome distribution.} Human review distribution in the reviewed prior authorization cases}
\label{tab:A4_reviewer_outcomes}
\begin{tabular}{l r}
\toprule
\textbf{Review Outcome} & \textbf{Frequency (\%)} \\
\midrule
Correct approval        & 2{,}901 (40.5\%) \\
False approval          & 230 (3.2\%) \\
Correct escalation      & 3{,}832 (53.5\%) \\
Unnecessary escalation  & 203 (2.8\%) \\
\bottomrule
\end{tabular}
\end{table}

\subsection{Sensitivity Analyses}
\label{App:B}

Table \ref{tab:A4_sensitivity_single_factor} reports single-factor adjusted analyses described in Section \ref{section:results_sensitivity}, in which the protected attribute is evaluated under one adjustment variable at a time. Each column corresponds to a separate logistic regression model including the protected attribute and only one control variable. These analyses are intended as diagnostic sensitivity checks and are not used for final fairness determinations. Datapoints with ``Unknown'' values for the protected attributes were excluded from this analysis. Results are interpreted using the $\delta=\pm$5pp tolerance band.

Single-factor adjusted estimates are broadly consistent with the fully adjusted, protocol-controlled results reported in the main text. For sex, age, and SES, estimated error-rate differences remain small and within the $\delta=\pm$5pp tolerance band across specifications, indicating that primary conclusions are not driven by any single adjustment variable. For race/ethnicity (and in a few other instances in other attributes), confidence intervals are wider and evidentiary strength is weaker across models, reflecting limited subgroup representation rather than systematic directional effects. These analyses are diagnostic and supplementary; final fairness conclusions are based on the protocol-controlled adjusted analyses reported in the main text.

\begin{table}[!ht]
\centering
\caption{ $|$ \textbf{Sensitivity analysis.} Single-Factor adjusted error-Rate disparities across protected attributes after controlling for demographic attributes or protocol. The table reports adjusted risk differences (percentage points), 95\% bootstrap confidence intervals (CI), inclusion within the predefined operational tolerance band ($\delta=\pm$5pp), the Evidence Ratio (CI width relative to $\delta$), and qualitative evidence strength. Across race, sex, age, and SES, adjusted disparities are mostly small and contained within the tolerance threshold with strong to moderate precision; on a few occasions, we obtain inconclusive results}

\label{tab:A4_sensitivity_single_factor}

\begin{tabular}{l l c l c l}
\toprule
\textbf{Attr.} & \textbf{Adj. Set} & \textbf{Adj.} $\boldsymbol{\Delta}$ & \textbf{95\% CI} & \textbf{CI $\boldsymbol{\in \delta}$} & \textbf{Evi. Ratio (Strength)} \\
\midrule
\textbf{Sex}
  & Protocol & $+0.49$ & $[-0.69, 1.66]$ & Yes & 23\% (Strong evidence) \\
  & Race     & $+0.40$ & $[-2.93, 3.67]$ & Yes & 66\% (Moderate evidence) \\
  & Age      & $+0.61$ & $[-0.63, 1.88]$ & Yes & 25\% (Strong evidence) \\
  & SES      & $+0.80$ & $[-1.16, 2.96]$ & Yes & 41\% (Strong evidence) \\[4pt]

\textbf{Race}
  & Protocol & $+0.18$ & $[-5.37, 4.80]$  & No & 102\% (Weak evidence) \\
  & Sex      & $+1.04$ & $[-3.22, 4.72]$  & Yes & 79\% (Moderate evidence) \\
  & Age      & $+0.28$ & $[-3.54, 4.50]$  & Yes & 80\% (Weak evidence) \\
  & SES      & $-4.24$ & $[-12.16, 2.55]$ & No & 147\% (Weak evidence) \\[4pt]

\textbf{Age}
  & Protocol & $+0.35$ & $[-0.83, 1.60]$ & Yes & 24\% (Strong evidence) \\
  & Sex      & $+0.26$ & $[-1.03, 1.49]$ & Yes & 25\% (Strong evidence) \\
  & Race     & $+2.16$ & $[-1.12, 5.51]$ & No & 66\% (Moderate evidence) \\
  & SES      & $-0.01$ & $[-2.18, 1.96]$ & Yes & 41\% (Strong evidence) \\[4pt]

\textbf{SES}
  & Protocol & $-0.32$ & $[-2.97, 2.52]$ & Yes & 55\% (Moderate evidence) \\
  & Sex      & $-0.23$ & $[-2.78, 2.29]$ & Yes & 51\% (Moderate evidence) \\
  & Race     & $+0.39$ & $[-6.41, 8.92]$ & No & 153\% (Weak evidence) \\
  & Age      & $+0.21$ & $[-2.46, 3.10]$ & Yes & 56\% (Moderate evidence) \\

\bottomrule
\end{tabular}
\end{table}

\subsection{Robustness to Missingness Handling in Adjusted Analyses}
\label{App:C}

Table \ref{tab:A5_robustness_complete_case} reports adjusted error-rate disparities from complete-case analyses, in which rows with missing covariate values (encoded as ``Unknown'') are excluded. Each column corresponds to a logistic regression model adjusting for one demographic covariate at a time, in addition to protocol, as described in Section \ref{section:methods_adjusted}. These analyses are included to assess the robustness of the primary adjusted findings to alternative approaches to handling missing demographic information.

Across sex, age, and SES, complete-case-adjusted estimates are consistent in magnitude and in their tolerance-band interpretation with the primary analyses that include ``Unknown'' covariate levels in Section \ref{section:results_adjusted}, indicating that fairness conclusions for these attributes are robust to missingness handling. For race/ethnicity (and, in some instances, when race is used as a covariate), complete-case analyses exhibit greater variability and wider confidence intervals due to further reductions in effective sample size, underscoring data limitations rather than substantive changes in estimated disparities. As with other supplementary analyses, these results do not alter the primary fairness conclusions reported in the main text.

\begin{table}[!ht]
\centering
\caption{ $|$ \textbf{Adjusted evaluation with ``Unknown''.} Adjusted error-rate disparities across protected attributes after controlling for protocol and additional demographic covariates (Complete-case covariates; no ``Unknown'' levels). The table reports adjusted risk differences (percentage points), 95\% bootstrap confidence intervals (CI), inclusion within the predefined operational tolerance band ($\delta=\pm$5pp), the Evidence Ratio (CI width relative to $\delta$), and qualitative evidence strength. Across sex, age, and SES, adjusted disparities are small and fully contained within the tolerance threshold with strong to moderate precision; race-adjusted comparisons (and cases where we use race as covariates) exhibit wider intervals due to limited sample size and are therefore less conclusive}

\label{tab:A5_robustness_complete_case}

\begin{tabular}{l l c l c l}
\toprule
\textbf{Attr.} & \textbf{Adj. Set} & \textbf{Adj.} $\boldsymbol{\Delta}$ & \textbf{95\% CI} & \textbf{CI $\boldsymbol{\in \delta}$} & \textbf{Evi. Ratio (Strength)} \\
\midrule
\textbf{Sex}
  & Protocol + Race & $+0.23$ & $[-3.30, 3.94]$ & Yes & 72\% (Moderate evidence) \\
  & Protocol + Age  & $+0.58$ & $[-0.60, 1.90]$ & Yes & 25\% (Strong evidence) \\
  & Protocol + SES  & $+0.58$ & $[-1.71, 3.00]$ & Yes & 47\% (Strong evidence) \\[4pt]

\textbf{Race}
  & Protocol + Sex  & $+1.00$ & $[-4.53, 5.19]$  & No & 97\% (Weak evidence) \\
  & Protocol + Age  & $+0.12$ & $[-5.38, 5.08]$  & No & 105\% (Weak evidence) \\
  & Protocol + SES  & $-3.91$ & $[-12.70, 4.18]$ & No & 169\% (Weak evidence) \\[4pt]

\textbf{Age}
  & Protocol + Sex  & $+0.02$ & $[-1.19, 1.21]$ & Yes & 24\% (Strong evidence) \\
  & Protocol + Race & $+1.18$ & $[-3.02, 5.47]$ & No & 85\% (Weak evidence) \\
  & Protocol + SES  & $-0.82$ & $[-3.61, 1.56]$ & Yes & 52\% (Moderate evidence) \\[4pt]

\textbf{SES}
  & Protocol + Sex  & $-0.85$ & $[-3.70, 2.02]$ & Yes & 57\% (Moderate evidence) \\
  & Protocol + Race & $+0.35$ & $[-8.40, 12.65]$ & No & 211\% (Weak evidence) \\
  & Protocol + Age  & $-0.57$ & $[-2.76, 2.42]$ & Yes & 52\% (Moderate evidence) \\

\bottomrule
\end{tabular}
\end{table}

\end{document}